\documentclass{article}

\usepackage[utf8]{inputenc}
\usepackage[T1]{fontenc}
\usepackage{lmodern}

\usepackage[margin=1in]{geometry}
\usepackage{microtype}

\usepackage{amsmath,amssymb}
\usepackage{bbm}
\usepackage{mathtools}
\usepackage{marginnote}
\usepackage[pdftex]{graphicx}

\usepackage{graphicx}
\usepackage{booktabs}
\usepackage{array}
\usepackage{tabularx}
\newcolumntype{C}{>{\centering\arraybackslash}X}
\usepackage{float}

\PassOptionsToPackage{hidelinks}{hyperref}
\usepackage[numbers,sort&compress]{natbib}
\usepackage{doi}
\DeclareUnicodeCharacter{2013}{--} 
\DeclareUnicodeCharacter{2014}{---}
\DeclareUnicodeCharacter{2019}{'}  

\graphicspath{{Figs/}{Figs/}}

\title{XBusNet: Text-Guided Breast Ultrasound Segmentation via
Multimodal Vision--Language Learning}

\author{%
  \href{https://orcid.org/0009-0002-6750-5604}{\includegraphics[scale=0.06]{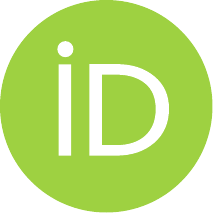}\hspace{1mm}Raja Mallina} \\
  Department of Computer Science\\
  University of Nevada, Las Vegas, Nevada, \\United States
  \and
  \href{https://orcid.org/0009-0006-2124-1032}{\includegraphics[scale=0.06]{orcid.pdf}\hspace{1mm}Bryar Shareef}\thanks{Corresponding author: bryar.shareef@unlv.edu}\\
  Department of Computer Science\\
  University of Nevada, Las Vegas, Nevada, \\United States
}

\date{} 
\begin{document}
\maketitle
\begin{abstract}
\textbf{Background:} Precise breast ultrasound (BUS) segmentation supports reliable measurement, quantitative analysis, and downstream classification, yet remains difficult for small or low-contrast lesions with fuzzy margins and speckle noise. Text prompts can add clinical context, but directly applying weakly localized text–image cues (e.g., CAM/CLIP-derived signals) tends to produce coarse, blob-like responses that smear boundaries unless additional mechanisms recover fine edges. \textbf{Methods:} We propose XBusNet, a \emph{novel dual-prompt, dual-branch multimodal model} that combines image features with clinically grounded text. A global pathway based on a CLIP Vision Transformer encodes whole-image semantics conditioned on lesion size and location, while a local U-Net pathway emphasizes precise boundaries and is modulated by prompts that describe shape, margin, and Breast Imaging Reporting and Data System (BI\mbox{-}RADS) terms. Prompts are assembled automatically from structured metadata, requiring no manual clicks. We evaluate on the Breast Lesions USG (BLU) dataset using five-fold cross-validation. Primary metrics are Dice and Intersection over Union (IoU); we also conduct size-stratified analyses and ablations to assess the roles of the global and local paths and the text-driven modulation. \textbf{Results:} XBusNet achieves state-of-the-art performance on BLU, with mean Dice of 0.8765 and IoU of 0.8149, outperforming six strong baselines. Small lesions show the largest gains, with fewer missed regions and fewer spurious activations. Ablation studies show complementary contributions of global context, local boundary modeling, and prompt-based modulation. \textbf{Conclusions:} A dual-prompt, dual-branch multimodal design that merges global semantics with local precision yields accurate BUS segmentation masks and improves robustness for small, low-contrast lesions.
\end{abstract}


\section{Introduction}
\label{sec:introduction}
Breast cancer is the most common cancer in U.S. women (about 30\% of new cases in 2025), and roughly 42{,}170 women are expected to die from it \cite{ACS2025KeyStats}. Early detection and timely treatment remain central to lowering mortality \cite{ACS2025KeyStats}.
Mammography, magnetic resonance imaging (MRI), and ultrasound are the main imaging modalities used for screening and diagnostic workups \cite{MammoLimitations,MRIReview}. Mammography and MRI are accurate, but each has limitations: mammography loses sensitivity in women with dense breast tissue \cite{MammoDenseBreast}, and MRI, though highly sensitive, is costly and impractical for routine use \cite{MRIcost}.
Ultrasound is safe, painless, nonionizing, affordable, and widely available, which is valuable for younger patients and in settings with limited resources \cite{BUSadvantages}. Its interpretation, however, is highly operator dependent and is often hindered by speckle noise, heterogeneous tissue appearance, and indistinct lesion boundaries, challenges shown in benchmark datasets such as BUSIS \cite{BUSIS, clsfBenchmark}.
These factors motivate automated segmentation methods that are accurate, robust across imaging conditions, and interpretable to radiologists in routine practice.

Accurate tumor segmentation improves diagnostic precision by enabling measurements aligned with BI\mbox{-}RADS (size, shape, and margins), yielding more reliable radiomics features, keeping classifiers focused on the lesion, and supporting longitudinal tracking \cite{BIRADSstandard,RadiomicsBUS}. Over the past decade, breast ultrasound (BUS) segmentation has progressed in three broad stages. Early studies used classical image processing—thresholding, edge detection, region growing, and active contours \cite{ClassicalBUS2}. Next came machine-learning pipelines that relied on hand-crafted texture and shape descriptors (e.g., GLCM, local binary patterns, wavelets) paired with supervised classifiers such as SVMs and random forests, or with clustering and graph-based methods (fuzzy c-means, k-means, conditional random fields, graph cuts) \cite{BUSIS}. With deep learning, UNet and related encoder–decoder models became the standard \cite{UNet,UNetDerivatives}. Variants that add residual or dense connections and attention gates improved boundary quality \cite{ResNet,DenseNet,AttentionUNet}. Attention mechanisms and Transformer designs helped capture long-range context \cite{AttentionReview,ViT,TransUNet}, and hybrid CNN–Transformer models reported strong results on BUS data, including a multitask (segmentation + classification) CNN–Transformer trained jointly for both tasks \cite{HybridKuan,hybridShareef}.
Despite this progress, BUS still presents challenging scenarios: small tumor sizes, low contrast, heterogeneous backgrounds, and variability across scanners and sites. These difficulties are compounded by limited labeled data, class imbalance, and annotation variability. Pixel only training also rarely uses explicit clinical descriptors (e.g., BI\mbox{-}RADS or radiomics), which can make outputs harder to interpret in routine reading \cite{BUSIS,VisionLimitations}. Small tumor–aware architectures such as STAN and ESTAN directly target this failure mode \cite{shareef2020stan,shareef2022estan}.

To address this gap, prior studies have injected domain knowledge into BUS segmentation in several ways. First, clinical descriptors and radiomics are used as auxiliary supervision or fused with visual features \cite{RadiomicsBUS,BIRADSuse,BIRADSFusion}. Second, anatomy-aware models and losses encode priors on shape, boundaries, and topology to keep masks plausible \cite{Oktay2017ACNN,BoundaryLoss,HausdorffLoss,TopoLoss,PriorLossSurvey}. Third, ultrasound physics has been leveraged through speckle and attenuation models and quantitative ultrasound features \cite{Yu2002SRAD,NakagamiReview,NakagamiBreast2023,QUSReview2024}. These strategies improve clinical grounding but still face limits in scalability and generalization, which motivates frameworks that flexibly integrate descriptors while preserving robustness across imaging conditions.

Prompt-based learning treats segmentation as a conditional task steered by auxiliary inputs (“prompts”). In practice, prompts take several forms:
\textit{(i) Spatial prompts.} Image coordinates or regions such as points, bounding boxes, polygons/masks, or scribbles. The Segment Anything Model (SAM) is a canonical spatially promptable segmenter and supports these types at scale \cite{SAM}. \textit{(ii) Textual prompts.} Class names or free-form phrases that describe the target. CLIPSeg shows that text or image prompts can condition masks \cite{CLIPSeg}. Pairing open-vocabulary detectors or vision–language models with SAM enables text-conditioned localization and segmentation; for example, Grounding DINO with SAM (“Grounded-SAM”) uses free text to propose regions, and BLIP provides strong image–text pretraining \cite{GroundingDINO,GroundedSAM,BLIP,COCO,RefCOCO}. \textit{(iii) Other side information.} Categorical tags or numeric attributes (e.g., approximate lesion size or laterality) that can be encoded as tokens or embeddings and used alongside visual features.

Adapting these ideas to medical imaging is challenging because lesions often have subtle contrast, boundaries are weak, and domain shift across scanners and sites is common \cite{SAMmedLimits}. Several adaptations target these issues: BUSSAM introduces CNN–ViT adapters for breast ultrasound, MedSAM pretrains on large medical datasets, and SAMed applies low-rank adaptation for efficient tuning \cite{BUSSAM,MedSAM,SAMed}. Vision–language approaches have also been explored directly for ultrasound, such as CLIP-TNseg\cite{CLIPTNseg}; causal interventions have been proposed to stabilize text guidance under noisy or ambiguous prompts \cite{CLIPTNseg,CausalCLIPSeg}. These studies indicate benefits from both spatial and text prompts, yet most methods adopt a single prompt modality. Few approaches jointly exploit global semantic context and fine-grained clinical attributes in a manner consistent with radiologist reasoning. Furthermore, it is desirable that segmentation outputs are consistent with BI\mbox{-}RADS descriptors \cite{BIRADSstandard}. Common tools such as Grad\mbox{-}CAM and attention visualizations offer partial insight but often produce diffuse or inconsistent patterns that are difficult to map to clinical terms \cite{GradCAM,AttentionXAI,XAIlimits}. 

In this work, we present XBusNet, a dual branch, dual prompt vision and language framework for BUS segmentation. A global branch based on Contrastive Language Image Pretraining (CLIP) and a Vision Transformer (ViT) is conditioned by a Global Feature Context Prompt (GFCP) that encodes lesion size and centroid to provide scene level semantics. A local branch based on UNet with multi head self attention preserves boundary detail. The local decoder is modulated by a text conditioned scaling and shifting mechanism, Semantic Feature Adjustment (SFA), which is driven by a Local or Attribute Guided Prompt (LFP) that encodes shape, margin, and Breast Imaging Reporting and Data System (BI\mbox{-}RADS) terms. Prompts are assembled reproducibly from structured metadata, require no manual clicks, and integrate with standard training. We evaluate XBusNet on the BLU dataset using five-fold cross validation, include ablations to isolate the role of each component, analyze performance across lesion sizes, and provide Grad\mbox{-}CAM visualizations as qualitative views of model focus relative to BI\mbox{-}RADS descriptors.

\hspace{1cm}

\textbf{Our contributions are as follows:}
\begin{itemize}
\item We propose \textbf{XBusNet}, a \textbf{dual branch, dual prompt} vision and language segmentation architecture that combines a CLIP ViT global branch conditioned by GFCP with a UNet local branch with multi head self attention that is modulated via SFA driven by LFP.
\item We design a \textbf{reproducible prompt pipeline} that converts structured metadata into natural language prompts, including global prompts for lesion size and centroid and local prompts for shape, margin, and BI\mbox{-}RADS.
\item We introduce a \textbf{lightweight SFA mechanism} in the local decoder to inject attribute aware scaling and shifting, improving boundary focus while preserving fine detail.
\item We provide a \textbf{comprehensive evaluation} protocol with five-fold cross validation, size stratified analysis, component ablations, and Grad\mbox{-}CAM overlays used as qualitative visualizations of model focus relative to BI\mbox{-}RADS descriptors.
\end{itemize}

\section{Materials and Methods}
\subsection{Datasets}
We used the Breast Lesions USG (BLU) dataset \cite{pawlowska2024curated}, which contains 256 breast ultrasound scans from 256 patients.
Each scan includes expert verified annotations for benign and malignant lesions.
Four images with multiple annotated lesions (two tumors in the same image) were excluded because they represent a very small fraction of the dataset (4/256, \(\approx\)1.6\%), which would yield an imbalanced and hard-to-interpret evaluation for per-image metrics. The remaining 252 images include 154 benign and 98 malignant tumors.

The dataset provides pixel wise segmentation masks, patient level clinical attributes such as age, breast tissue composition, signs, and symptoms, image level BI\mbox{-}RADS labels, and tumor level diagnostic information such as shape, margin, and histopathology.
\begin{figure}[htbp]
  \centering
  \includegraphics[width=\textwidth]{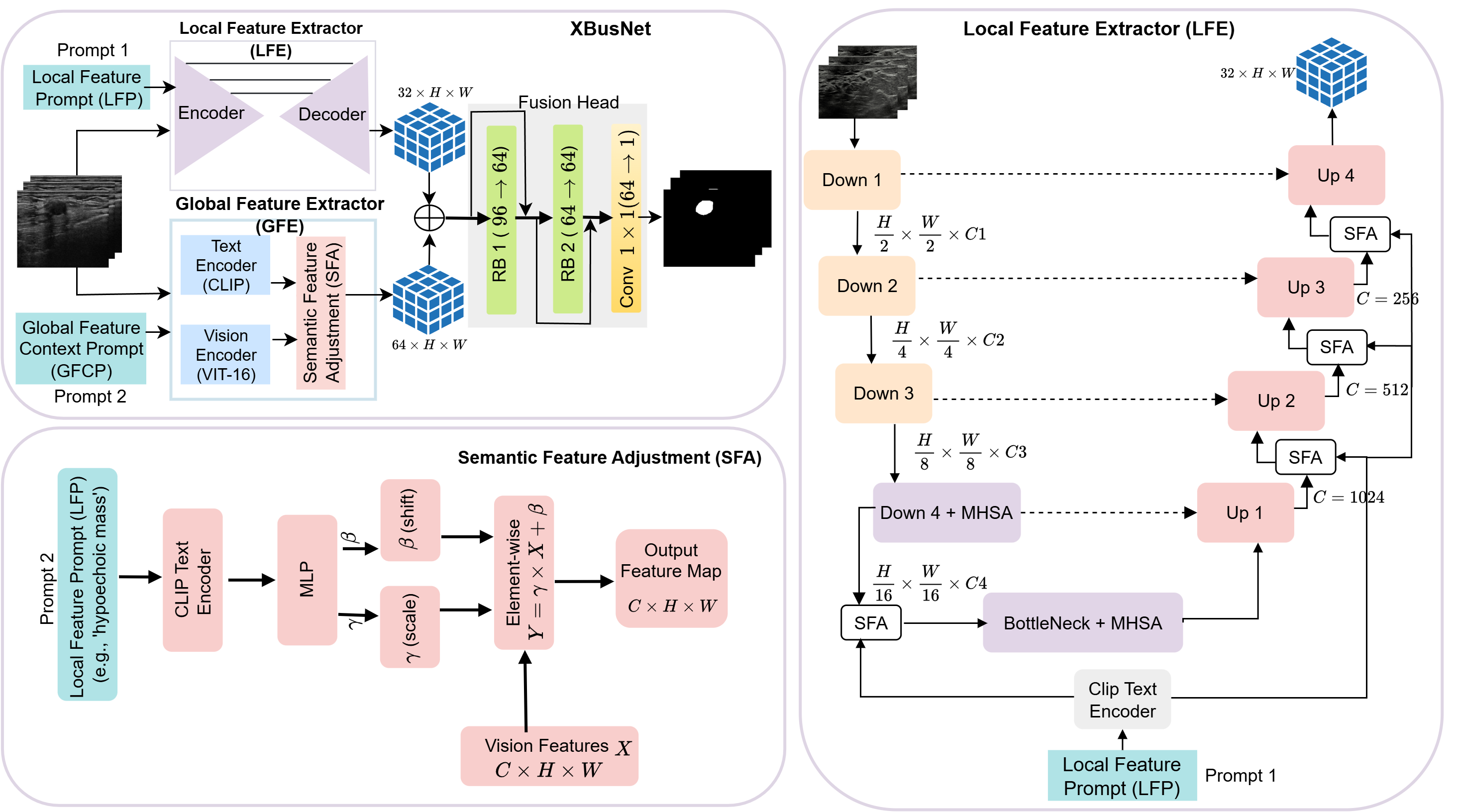} 
  \caption{Overview of the proposed XBusNet architecture, showing the Local Feature Extractor (LFE), Global Feature Extractor (GFE), and Semantic Feature Adjustment (SFA) modules. RB = Residual Block; MHSA = Multi-Head Self-Attention.}
  \label{fig:myimage}
\end{figure}
\subsection{Preprocessing and Prompt Construction}
All images were resized to $352 \times 352$ pixels. Prompts were derived from structured metadata in the dataset CSV.
The \emph{Global Feature Context Prompt} encodes lesion \emph{size} and \emph{location}.
Size is taken from the dataset metadata and discretized using training-set quantiles (small/medium/large).
\textbf{Location handling differs by split:}
for \emph{training images}, location is the centroid of the reference segmentation mask computed from image moments,
\begin{equation}
C_x=\frac{M_{10}}{M_{00}},\qquad C_y=\frac{M_{01}}{M_{00}},
\end{equation}
where $M_{00}$ is the mask area and $M_{10},M_{01}$ are spatial moments.
For \emph{validation/test images}, location is estimated \emph{without masks} from a first-pass probability map as detailed in Section~\ref{sec:gfe}.
The \emph{Local Feature Prompt} encodes clinically used attributes—\emph{shape}, \emph{margin}, and \emph{BI\mbox{-}RADS}—verbalized directly from metadata.
Prompts are tokenized and embedded with the CLIP text encoder as described in the model subsections.
\label{sec:prompt_construction}

\subsection{Training Setup}\label{sec:train}
We used five-fold cross validation.
For each fold we split the 252 annotated images into train and validation sets with an 80 to 20 ratio, ensuring non overlapping images.
Unless stated otherwise, training ran for 1000 iterations per fold with batch size 4, the AdamW optimizer, an initial learning rate of $10^{-4}$, and a cosine schedule.
Mixed precision training was enabled with automatic mixed precision.

\subsection{Evaluation Metrics}\label{sec:metrics}
We report Dice, Intersection over Union (IoU), false positive rate (FPR), and false negative rate (FNR).
Let $M \in \{0,1\}^{H\times W}$ be the ground-truth mask and $P \in [0,1]^{H\times W}$ the predicted probability map.
We obtain a binary prediction by thresholding at $\tau_{\mathrm{seg}}{=}0.5$:
\[
\hat{Y}=\mathbbm{1}\!\left\{\,P \ge \tau_{\mathrm{seg}}\,\right\} \in \{0,1\}^{H\times W}.
\]

Pixel-wise counts (sums over all pixels $(i,j)$) are
\[
\mathrm{TP}=\sum_{i,j}\hat{Y}_{ij} M_{ij},\quad
\mathrm{FP}=\sum_{i,j}\hat{Y}_{ij} (1-M_{ij}),\quad
\mathrm{FN}=\sum_{i,j}(1-\hat{Y}_{ij}) M_{ij},\quad
\mathrm{TN}=\sum_{i,j}(1-\hat{Y}_{ij})(1-M_{ij}).
\]
Per-image metrics are
\[
\mathrm{Dice}=\frac{2\,\mathrm{TP}}{2\,\mathrm{TP}+\mathrm{FP}+\mathrm{FN}},\qquad
\mathrm{IoU}=\frac{\mathrm{TP}}{\mathrm{TP}+\mathrm{FP}+\mathrm{FN}},
\]
\[
\mathrm{FNR}=\frac{\mathrm{FN}}{\mathrm{FN}+\mathrm{TP}},\qquad
\mathrm{FPR}=\frac{\mathrm{FP}}{\mathrm{FP}+\mathrm{TN}}.
\]
We compute metrics per image and average within each fold; fold means are then summarized across the $K{=}5$ folds (mean$\pm$SD).
For numerical stability, we use $\varepsilon=10^{-8}$ in denominators when implementing.

\subsection{Implementation Details and Hardware}\label{sec:impl}
All experiments were implemented in PyTorch and run on a cluster with $8 \times$ NVIDIA A100 Tensor Core GPUs in SXM form factor. The CLIP encoders are frozen during training. The ResNet50 encoder is initialized from ImageNet weights and fine tuned.

\subsection{Overall Architecture}
XBusNet is a text guided medical image segmentation framework that integrates fine grained local context from medical images with global semantic cues derived from natural language prompts.
The model follows a dual branch, dual prompt vision and language design with a Global Feature Extractor (GFE) and a Local Feature Extractor (LFE) that run in parallel and exchange information through a Semantic Feature Adjustment (SFA) mechanism.
The GFE uses a CLIP based Vision Transformer with patch size 16 to capture global semantic relationships conditioned on a high level conditional prompt $p_{c}$.
The LFE uses a ResNet50 encoder with transformer blocks at deep levels and a UNet style decoder to represent spatial detail under a phrase level prompt $p_{\ell}$.

Let $I \in \mathbb{R}^{B \times 3 \times H \times W}$ denote the input image batch and let $p_{c}$ and $p_{\ell}$ be the conditional and local prompts.
The two branches produce
\begin{equation}
    \mathbf{F}_{g} = \mathrm{GFE}(I, p_{c}), \qquad 
    \mathbf{F}_{\ell} = \mathrm{LFE}(I, p_{\ell}),
\end{equation}
with $\mathbf{F}_{g} \in \mathbb{R}^{B \times C_{g} \times h \times w}$ and $\mathbf{F}_{\ell} \in \mathbb{R}^{B \times C_{\ell} \times h \times w}$.
SFA modulates features in both branches using scaling and shifting parameters computed from the corresponding prompt embedding:
\begin{equation}
\begin{aligned}
\hat{\mathbf{F}}_{g} &= \boldsymbol{\gamma}_{g} \odot \mathbf{F}_{g} + \boldsymbol{\beta}_{g}, 
&\quad [\boldsymbol{\gamma}_{g}, \boldsymbol{\beta}_{g}] &= \mathrm{SFA}(\mathbf{F}_{g}, \mathbf{e}_{c}),\\
\hat{\mathbf{F}}_{\ell} &= \boldsymbol{\gamma}_{\ell} \odot \mathbf{F}_{\ell} + \boldsymbol{\beta}_{\ell}, 
&\quad [\boldsymbol{\gamma}_{\ell}, \boldsymbol{\beta}_{\ell}] &= \mathrm{SFA}(\mathbf{F}_{\ell}, \mathbf{e}_{\ell}),
\end{aligned}
\label{eq:sfa-both-branches}
\end{equation}
where $\odot$ denotes element wise multiplication.
A residual form can be used at implementation time as 
$\mathbf{F}_{\text{out}} = \hat{\mathbf{F}} + \mathbf{F}$.

The fusion head integrates the modulated features and prepares them for prediction:
\begin{equation}
\mathbf{F}_{\text{fused}} = \Phi_{\text{fuse}}\!\left([\hat{\mathbf{F}}_{g}, \hat{\mathbf{F}}_{\ell}]\right),
\label{eq:fusion}
\end{equation}
Here, $\Phi_{\mathrm{fuse}}$ denotes the two residual blocks described in Section~\ref{sec:fusion}.
This is followed by a $1{\times}1$ convolution that yields the segmentation logits:
\begin{equation}
    \hat{\mathbf{Y}} = \mathrm{Conv}_{1\times1}(\mathbf{F}_{\mathrm{fused}}), 
    \qquad \hat{\mathbf{Y}} \in \mathbb{R}^{B \times 1 \times H \times W}.
\end{equation}
This organization lets the network use language to steer scene-level context while the local pathway preserves boundaries and fine structure.

\subsection{Global Feature Extractor (GFE)}
\label{sec:gfe}

\paragraph{Global features and prompt.}
Global features comprise \emph{size} and \emph{location}. Lesion size is read from the dataset metadata and discretized using training-set quantiles (small/medium/large). 
\textbf{Location handling differs by split:} 
\emph{(i) Training)} the location is the centroid \((C_x,C_y)\) of the reference segmentation mask, computed from image moments \(C_x=M_{10}/M_{00}\), \(C_y=M_{01}/M_{00}\); 
\emph{(ii) Validation/Test)} the location is estimated from a first-pass probability map \(P\) by running the model without spatial text, thresholding at \(\tau=0.30\), retaining the largest connected component to form a coarse proposal \(R\), and computing its centroid \((\hat C_x,\hat C_y)\). 
Centroids are mapped to breast quadrants (upper/lower \(\times\) inner/outer) to produce a location token. 
Size and location tokens are verbalized to form the Global Feature Context Prompt \(p_c\). 
Quantiles and thresholds are fixed a priori per fold using only the corresponding training split.

\paragraph{GFE computation.}
The GFE computes a scene-level representation conditioned on \(p_c\) using a vision transformer. 
The prompt is embedded by the text encoder to obtain \(\mathbf{e}_{c}\in\mathbb{R}^{d}\).
Selected transformer layers provide token sequences \(\{\mathbf{Z}^{(j)}\}_{j\in\mathcal{L}}\) with \(\mathbf{Z}^{(j)}\in\mathbb{R}^{B\times N\times D}\).
Tokens are linearly reduced to a common width and aggregated,
\begin{equation}
\mathbf{A}=\sum_{j\in\mathcal{L}}\mathbf{W}^{(j)}\,\mathbf{Z}^{(j)}_{[:,\,1:\,,\,:]}\;\in\;\mathbb{R}^{B\times (N-1)\times r},
\end{equation}
where the class token is removed.
Prompt conditioning is applied as channel-wise scaling and shifting of the reduced tokens at a chosen transformer depth,
\begin{equation}
\mathbf{c}=\mathbf{W}_{c}\,\mathbf{e}_{c}, \qquad
\tilde{\mathbf{A}} = \big(\mathbf{W}_{\mathrm{mul}}\mathbf{c}\big)\odot \mathbf{A} + \big(\mathbf{W}_{\mathrm{add}}\mathbf{c}\big).
\end{equation}
The conditioned tokens are reshaped to a spatial grid and projected with a transposed convolution to obtain the global feature map
\begin{equation}
\mathbf{F}_{g}\in\mathbb{R}^{B\times 64\times h\times w}.
\end{equation}
SFA in the global branch follows the affine form in Eq.~\eqref{eq:sfa-core} and produces \(\hat{\mathbf{F}}_{g}\) as in Eq.~\eqref{eq:sfa-both-branches}.
The vision and text encoders are frozen; only the reduction, conditioning, and projection layers are trained.
\subsection{Local Feature Extractor (LFE)}
\label{sec:lfe}

\paragraph{Local features and prompt.}
Local features comprise \emph{shape}, \emph{margin}, and \emph{BI\mbox{-}RADS}. 
These attributes are read from the dataset metadata and verbalized to form the Local Feature Prompt \(p_\ell\) (e.g., ``irregular shape, microlobulated margin, BI\mbox{-}RADS 4''). 
The mapping from metadata to tokens is deterministic and applied identically across folds.

\paragraph{LFE computation.}
The LFE recovers boundaries and fine structure under guidance from \(p_{\ell}\).
We use a ResNet50 encoder with transformer blocks at deep levels and a U\hbox{-}Net–style decoder.
The prompt is embedded with the text encoder,
\begin{equation}
    \mathbf{e}_{\ell} = \mathcal{T}(p_{\ell}) \in \mathbb{R}^{d}.
\end{equation}
Given \(I \in \mathbb{R}^{B \times 3 \times H \times W}\), the encoder yields features \(\{\mathbf{enc1},\mathbf{enc2},\mathbf{enc3},\mathbf{enc4},\mathbf{enc5}\}\).
We add a transformer block at the fourth encoder stage and at the bottleneck,
\begin{equation}
\mathbf{enc4}'=\mathrm{TrEnc}(\mathbf{enc4}),\qquad
\mathbf{center}'=\mathrm{TrEnc}(\mathbf{enc5}).
\end{equation}
The decoder upsamples and merges skip connections using up blocks to produce \(\mathbf{dec4},\mathbf{dec3},\mathbf{dec2},\mathbf{dec1}\), followed by a final up projection to obtain a 32-channel map at the decoder output.

Semantic feature adjustment (SFA) uses the local prompt embedding at encoder stage four and after the first three decoder up blocks,
\begin{equation}
\begin{aligned}
\mathbf{enc4}^{\star} &= \mathrm{SFA}(\mathbf{enc4}',\, \mathbf{e}_{\ell}),\\
\mathbf{dec4}^{\star} &= \mathrm{SFA}(\mathbf{dec4},\, \mathbf{e}_{\ell}),\\
\mathbf{dec3}^{\star} &= \mathrm{SFA}(\mathbf{dec3},\, \mathbf{e}_{\ell}),\\
\mathbf{dec2}^{\star} &= \mathrm{SFA}(\mathbf{dec2},\, \mathbf{e}_{\ell}).
\end{aligned}
\end{equation}
The final local representation is taken after the last decoder stage,
\begin{equation}
\mathbf{F}_{\ell}\in\mathbb{R}^{B\times 32\times h\times w}.
\end{equation}
This configuration preserves boundary detail and small structures while injecting attribute cues at the stages listed above.
Unless stated otherwise, the text encoder is frozen, and the LFE encoder–decoder parameters are trainable.

\subsection{Semantic Feature Adjustment (SFA)}
SFA injects prompt driven semantics into feature maps through channel wise scaling and shifting.
The module takes a visual tensor and a prompt embedding, predicts modulation parameters, and applies an affine transformation with an optional residual.

Let $\mathbf{F}\in\mathbb{R}^{B\times C\times h\times w}$ be a feature map and let $\mathbf{e}\in\mathbb{R}^{d}$ be the embedding of the associated prompt.
The core operation is
\begin{equation}
\begin{aligned}
\hat{\mathbf{F}} &= \boldsymbol{\gamma}\odot \mathbf{F}+\boldsymbol{\beta},\\
\big[\boldsymbol{\gamma},\,\boldsymbol{\beta}\big] &= \Psi(\mathbf{e}),
\end{aligned}
\label{eq:sfa-core}
\end{equation}
where $\odot$ denotes element wise multiplication and $\boldsymbol{\gamma},\boldsymbol{\beta}\in\mathbb{R}^{B\times C\times 1\times 1}$ broadcast over spatial dimensions.
Parameters are produced by a small projection network for each stage $k$ with channel width $C^{(k)}$,
\begin{equation}
\begin{aligned}
\mathbf{z}^{(k)} &= \mathrm{MLP}^{(k)}(\mathbf{e})\in\mathbb{R}^{2C^{(k)}},\\
\big[\boldsymbol{\gamma}^{(k)},\,\boldsymbol{\beta}^{(k)}\big] &= \mathrm{split}\!\left(\mathbf{z}^{(k)}\right),
\end{aligned}
\end{equation}
and are broadcast over spatial dimensions.

In the global branch we condition on $\mathbf{e}_{c}$ to obtain $\hat{\mathbf{F}}_{g}$; in the local decoder we condition on $\mathbf{e}_{\ell}$ to obtain $\hat{\mathbf{F}}_{\ell}$, as summarized in Eq.~\eqref{eq:sfa-both-branches}.

An additive skip may be used in implementation,
\begin{equation}
\mathbf{F}_{\mathrm{out}}=\hat{\mathbf{F}}+\mathbf{F}.
\end{equation}

\subsection{Feature Fusion and Prediction Head}
\label{sec:fusion}
We concatenate the modulated global and local feature maps to obtain
\begin{equation}
\mathbf{F}_{\mathrm{cat}} = \hat{\mathbf{F}}_{g}\,\Vert\,\hat{\mathbf{F}}_{\ell}\;\in\;\mathbb{R}^{B\times 96\times h\times w},
\end{equation}
and refine with two residual blocks followed by a $1{\times}1$ convolution,
\begin{equation}
\mathbf{F}_{1}=\mathrm{RB}_{1}(\mathbf{F}_{\mathrm{cat}}),\qquad
\mathbf{F}_{\mathrm{fused}}=\mathrm{RB}_{2}(\mathbf{F}_{1}),\qquad
\hat{\mathbf{Y}}=\mathrm{Conv}_{1\times1}(\mathbf{F}_{\mathrm{fused}}).
\end{equation}

\section{Results}
We follow the training setup in Section~\ref{sec:train} and the metrics in Section~\ref{sec:metrics}. All experiments use the implementation and hardware in Section~\ref{sec:impl}.

\subsection{Overall Performance}
We compare XBusNet with representative \emph{non-prompt} baselines including UNet~\cite{UNet}, U\_ResNet, U\_KAN~\cite{Ukan}, and AnatoSegNet~\cite{xu2025anatosegnet}, as well as \emph{prompt-guided} baselines such as BUSSAM~\cite{BUSSAM} and CLIP\_TNseg~\cite{CLIPTNseg}. All models are evaluated on BLU under the same five-fold cross-validation protocol (Section~\ref{sec:train}), the same pre-processing (Section~\ref{sec:prompt_construction}), and a common operating threshold $\tau_{\mathrm{seg}}{=}0.5$ (Section~\ref{sec:metrics}). We report mean$\pm$SD across folds for Dice, IoU, \emph{FPR}, and \emph{FNR} in Table~\ref{tab:overall}, and provide a size-stratified analysis in Table~\ref{tab:sizebins}.

\textbf{Fold-wise performance.}
With five-fold cross-validation (Table~\ref{tab:foldwise_metrics}), XBusNet attains consistently strong overlap across folds (Dice $0.8583$–$0.8910$; IoU $0.7987$–$0.8302$) with stable FPR and FNR, indicating reliable behavior across splits.

\textbf{Comparison with prior methods.}
Table~\ref{tab:overall} summarizes mean cross-validation performance against U\mbox{-}Net variants, anatomy-aware models, and text-guided segmenters. XBusNet achieves the best Dice ($0.8766$) and IoU ($0.8150$), and the lowest FNR ($0.0775$), reflecting fewer missed lesion pixels. FPR is comparable to the text-guided baseline and higher than some convolutional baselines that tend to under-segment; this trade-off favors recovering lesion extent while keeping spurious activations controlled.

\begin{figure}[htbp]
  \centering
  \includegraphics[width=\textwidth]{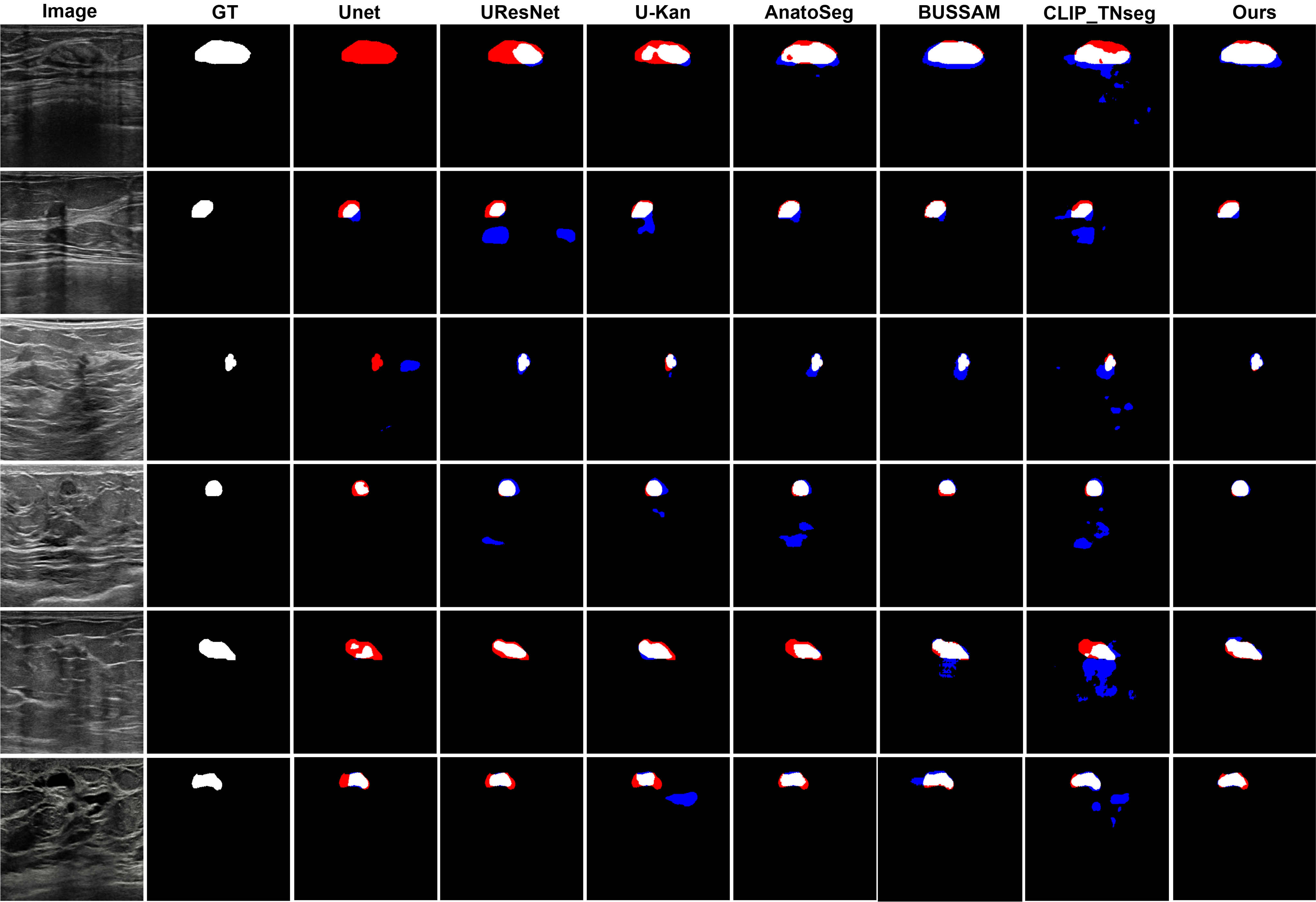}
  \caption{Qualitative comparison of breast ultrasound segmentations. False positives (FP) are shown in blue and false negatives (FN) in red.}
  \label{fig:qualitative_comparison}
\end{figure}

\textbf{Qualitative assessment.}
Figure~\ref{fig:qualitative_comparison} shows typical error modes with false positives (blue) and false negatives (red). First, for high-contrast masses, UNet variants exhibit small gaps at the posterior margin (missed pixels), whereas XBusNet produces continuous contours with fewer misses, consistent with the lower FNR in Table~\ref{tab:overall}. Second, on small or faint lesions, several baselines fragment the mask or miss the target entirely; XBusNet recovers a larger fraction of the lesion area, mirroring the gains in the 0--110\,px bin in Table~\ref{tab:sizebins}. Third, in low-contrast heterogeneous backgrounds, some text-guided baselines can add foreground in adjacent tissue, increasing FPR; XBusNet suppresses these spurious activations while retaining the faint rim. Finally, for irregular echotexture, convolutional baselines scatter foreground outside the mass, while XBusNet better follows the irregular edge.

\hspace{1cm}
\begin{table}
\centering
\caption{Overall comparison on BLU (five-fold CV). Values are mean$\pm$SD across folds at a fixed threshold $\tau_{\mathrm{seg}}{=}0.5$. Lower is better for FPR and FNR.}
\label{tab:overall}
\begin{tabularx}{\textwidth}{lCCCC}
\toprule
\textbf{Method} & \textbf{Dice (mean$\pm$SD)} & \textbf{IoU (mean$\pm$SD)} & \textbf{FPR (mean$\pm$SD)} & \textbf{FNR (mean$\pm$SD)} \\
\midrule
UNet~\cite{UNet}                 & 0.604$\pm$0.038 & 0.500$\pm$0.032 & 0.019$\pm$0.015 & 0.342$\pm$0.069 \\
U\mbox{-}ResNet                  & 0.721$\pm$0.040 & 0.621$\pm$0.049 & 0.010$\pm$0.007 & 0.264$\pm$0.078 \\
U\mbox{-}KAN~\cite{Ukan}         & 0.752$\pm$0.028 & 0.614$\pm$0.037 & \textbf{0.009$\pm$0.001} & 0.305$\pm$0.036 \\
AnatoSegNet~\cite{xu2025anatosegnet} & 0.733$\pm$0.072 & 0.627$\pm$0.084 & 0.015$\pm$0.013 & 0.216$\pm$0.034 \\
BUSSAM~\cite{BUSSAM}             & 0.833$\pm$0.020 & 0.735$\pm$0.026 & 0.014$\pm$0.003 & 0.101$\pm$0.024 \\
CLIP\mbox{-}TNseg~\cite{CLIPTNseg} & 0.839$\pm$0.019 & 0.765$\pm$0.020 & 0.189$\pm$0.002 & 0.134$\pm$0.017 \\
XBusNet (Ours)                   & \textbf{0.877$\pm$0.014} & \textbf{0.815$\pm$0.013} & 0.099$\pm$0.020 & \textbf{0.078$\pm$0.012} \\
\bottomrule
\end{tabularx}
\end{table}

\begin{table}
\centering
\caption{Dice and IoU scores across different tumor length intervals.}
\label{tab:sizebins}
\begin{tabular}{lcccccc}
\toprule
\textbf{Model / Length} & \multicolumn{2}{c}{\textbf{0--110}} & \multicolumn{2}{c}{\textbf{111--250}} & \multicolumn{2}{c}{\textbf{250+}} \\
\cmidrule(lr){2-3}\cmidrule(lr){4-5}\cmidrule(lr){6-7}
 & Dice & IoU & Dice & IoU & Dice & IoU \\
\midrule
UNet\cite{UNet}              & 0.3668 & 0.2750 & 0.6461 & 0.5251 & 0.6739 & 0.5625 \\
UResNet                      & 0.6238 & 0.5327 & 0.7859 & 0.6863 & 0.7305 & 0.6217 \\
U-Kan\cite{Ukan}            & 0.7256 & 0.5817 & 0.7584 & 0.6216 & 0.7312 & 0.5892 \\
AnatoSegNet\cite{xu2025anatosegnet}  & 0.6303 & 0.5215 & 0.7843 & 0.6822 & 0.7207 & 0.6131 \\
BUSSAM\cite{BUSSAM}         & 0.7846 & 0.6714 & 0.8465 & 0.7510 & 0.8426 & 0.7495 \\
CLIP-TNseg\cite{CLIPTNseg}  & 0.7689 & 0.7026 & 0.8587 & 0.7867 & 0.8447 & 0.7621 \\
XBusNet (Ours)              & \textbf{0.8507} & \textbf{0.7925} & \textbf{0.8947} & \textbf{0.8388} & \textbf{0.8553} & \textbf{0.7774} \\
\bottomrule
\end{tabular}
\end{table}

\hspace{1cm}

\begin{table}[ht]
\centering
\caption{Fold-wise validation performance of XBusNet (latest run).}
\label{tab:foldwise_metrics}
\begin{tabular}{lcccc}
\toprule
\textbf{Fold} & \textbf{Dice} & \textbf{IoU} & \textbf{FPR} & \textbf{FNR} \\
\midrule
0 & 0.8846 & 0.8241 & 0.0984 & 0.0670 \\
1 & 0.8910 & 0.8302 & 0.1280 & 0.0659 \\
2 & 0.8583 & 0.7987 & 0.1077 & 0.0835 \\
3 & 0.8649 & 0.8033 & 0.0857 & 0.0771 \\
4 & 0.8836 & 0.8181 & 0.0771 & 0.0944 \\
\textbf{Mean} & \textbf{0.8765} & \textbf{0.8149} & \textbf{0.0994} & \textbf{0.0776} \\
\bottomrule
\end{tabular}
\end{table}

\noindent\textbf{Statistical comparison.} Paired Wilcoxon signed-rank tests on image-level Dice and IoU showed that XBusNet significantly outperformed all baselines (two-sided $p<10^{-4}$ for both metrics across all pairwise tests). Effect sizes using the rank-biserial correlation indicated large advantages for XBusNet (median $|r_{rb}|\approx 0.82$ for Dice; median $|r_{rb}|\approx 0.80$ for IoU).

\subsection{Ablation Study}
We quantify the contribution of the Local Feature Extractor (LFE), the Global Feature Extractor (GFE), and Semantic Feature Adjustment (SFA) by toggling each component. Table~\ref{tab:ablation} shows that removing any single component degrades performance. Without LFE, Dice falls from $0.8766$ to $0.8572$ and IoU from $0.8150$ to $0.7865$, underscoring the role of local detail and boundary focus. Without GFE, Dice drops to $0.8453$ and IoU to $0.7772$, highlighting the value of global semantic cues from the conditional prompt. Without SFA, Dice decreases to $0.8600$ and IoU to $0.8068$, showing the benefit of prompt-conditioned modulation that aligns visual features with clinical attributes. The full model with all three components yields the best results, indicating complementary effects.

\begin{table}[htbp]
    \centering
    \caption{Ablation study on the effect of LFE, GFE, and SFA modules}
    \label{tab:ablation}
    \begin{tabular}{ccc|cc}
        \hline
        \textbf{LFE} & \textbf{GFE} & \textbf{SFA} & \textbf{Dice} & \textbf{IoU} \\
        \hline
        Yes & Yes & Yes & \textbf{0.8765} & \textbf{0.8149} \\
        No  & Yes & No  & 0.8572 & 0.7865 \\
        Yes & No  & Yes & 0.8453 & 0.7772 \\
        Yes & Yes & No  & 0.8600 & 0.8068 \\
        \hline
    \end{tabular}
\end{table}

\subsection{Qualitative Visualizations}
\begin{figure}[h]
  \centering
  \includegraphics[width=\textwidth]{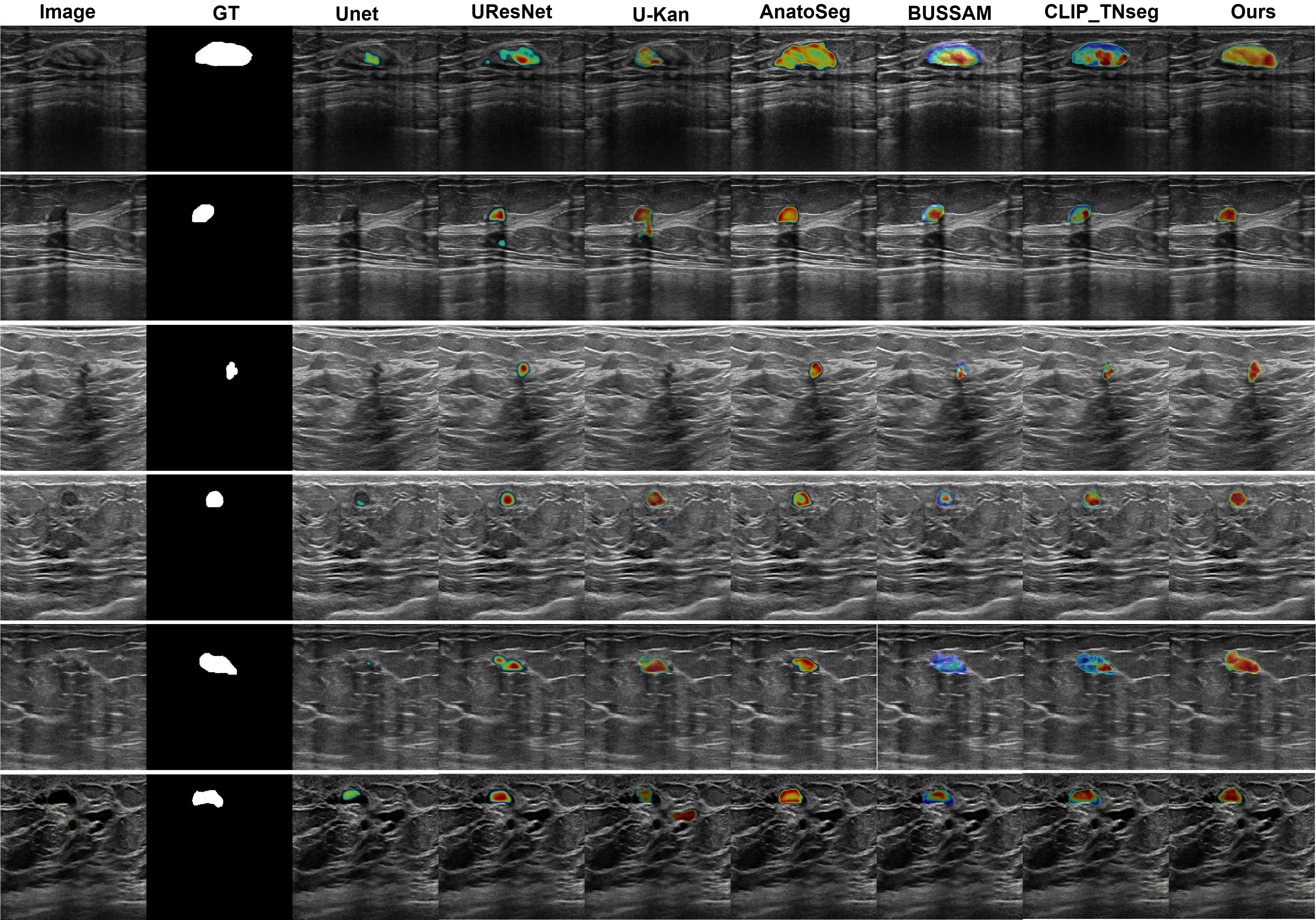}
  \caption{Grad-CAM comparison.}
  \label{fig:gradcam}
\end{figure}

We employ Grad\mbox{-}CAM overlays to visualize image regions that most influence the predicted masks (Figure~\ref{fig:gradcam}). Among the compared methods, UNet~\cite{UNet}, U\mbox{-}ResNet, U\mbox{-}KAN~\cite{Ukan}, and AnatoSegNet~\cite{xu2025anatosegnet} are non-prompt methods; BUSSAM~\cite{BUSSAM} and CLIP\mbox{-}TNseg~\cite{CLIPTNseg} are prompt-guided; XBusNet uses dual prompts (global and attribute-conditioned). Non-prompt baselines frequently display diffuse responses in low-contrast tissue, particularly posterior to the lesion, corresponding to missed or fragmented masks. Prompt-guided methods improve spatial correspondence; CLIP\mbox{-}TNseg may still exhibit scattered hotspots around speckle, whereas XBusNet shows concentrated activation adjacent to lesion boundaries and within the mass. For very small lesions, the salient regions for XBusNet overlap the recovered mask more consistently, in line with the small-lesion gains reported in Table~\ref{tab:sizebins}. These visualizations are qualitative and are not treated as formal explanations.

\section{Discussion}
XBusNet showed strong performance on BLU. Across five-folds, the model achieved a mean Dice of 0.8765 and an IoU of 0.8149, with narrow fold-to-fold variation, which suggests stable behavior under fold-wise splits. Compared with UNet variants, anatomy-aware models, and text-guided baselines, XBusNet also produced a lower false negative rate, indicating fewer missed lesion pixels. The false positive rate was comparable to text-guided baselines but higher than some convolutional models that under-segment; this trade-off favors recovering lesion extent and can be tuned by adjusting the operating threshold when a stricter specificity is required.

Size-stratified analysis supports these observations. The largest relative gains appear for the smallest lesions, where the model maintains competitive Dice and IoU, while performance remains strong for medium and large lesions. This pattern is consistent with the design: global prompt cues (size and approximate location) guide the search toward plausible regions, and local attribute cues (shape, margin, BI\mbox{-}RADS) help the decoder maintain sharp boundaries when pixel evidence is weak.

Ablation results point to complementary roles for both branches and for semantic feature adjustment. Removing the local branch reduces Dice and IoU, highlighting the importance of boundary detail. Disabling the global branch leads to a larger drop, showing that scene-level cues improve localization and reduce misses. Turning off semantic feature adjustment also degrades performance, which indicates that prompt-conditioned scaling and shifting adds value beyond the two-branch backbone. Together, these findings support the view that the full configuration is needed to realize the observed gains.

Qualitative panels mirror these trends. On high-contrast masses, several baselines show small gaps at the posterior margin, whereas XBusNet yields continuous contours, in line with the lower false negative rate. For small or faint lesions, baselines may fragment or miss the target, while XBusNet recovers a larger fraction of the lesion area. In heterogeneous backgrounds, some text-guided baselines spill into adjacent tissue; XBusNet restrains these activations while preserving the rim. With irregular echotexture, convolutional baselines scatter foreground outside the mass, whereas XBusNet follows the irregular edge more closely. Grad\mbox{-}CAM overlays are used as qualitative views of model focus and are not treated as formal explanations.

This study is limited by the availability of suitable public data. To our knowledge, there are no widely available breast ultrasound datasets that jointly provide pixel-level masks \emph{and} rich clinical descriptors such as BI\mbox{-}RADS terms, structured lesion metadata (e.g., shape and margin), and histopathological outcomes. This is the main reason we did not combine BLU with a public external cohort for evaluation. As a practical recommendation for future dataset releases, we encourage curators to include not only classification labels and segmentation masks, but also standardized BI\mbox{-}RADS descriptors, acquisition metadata, radiology reports, and, when possible, histopathology results. Such completeness would enable rigorous text-guided segmentation research, cross-dataset evaluation, and clinically meaningful analyses.

\section*{Conclusions}
We introduced XBusNet, a dual branch, dual prompt model for breast ultrasound segmentation that merges global context with local attribute guidance through semantic feature adjustment. On the BLU dataset, the method delivered strong segmentation performance with consistent gains on small lesions, reducing misses and spurious responses. The ablations show that the global branch, the local branch, and the modulation each contribute to the overall improvement, with the combined design offering balanced boundary quality and region coverage. These findings suggest that simple, automatically assembled text cues can strengthen ultrasound segmentation without changes to clinical imaging practice, and that coupling scene context with attribute prompts is a practical recipe for small target cases. 
 Future work will test cross center generalization, add boundary oriented metrics and calibration analysis, and explore coupling with detection and structured reporting.
 
\section*{Author Contributions}{Conceptualization, Bryar Shareef and Raja Mallina; methodology, Bryar Shareef and Raja Mallina; software, Raja Mallina; validation, Bryar Shareef and Raja Mallina; formal analysis, Bryar Shareef and Raja Mallina; investigation, Raja Mallina; resources, Bryar Shareef; data curation, Raja Mallina; writing—original draft preparation, Raja Mallina and Bryar Shareef; writing—review and editing, Bryar Shareef and Raja Mallina; visualization, Raja Mallina; supervision, Bryar Shareef; project administration, Bryar Shareef; funding acquisition, Not applicable. All authors have read and agreed to the published version of the manuscript.}

\newpage
\bibliographystyle{unsrtnat}
\bibliography{main}
\end{document}